\documentclass[lettersize,journal]{IEEEtran}
\usepackage{amsmath,amsfonts}
\usepackage{algorithmic}
\usepackage{algorithm}
\usepackage{array}
\usepackage[caption=false,font=footnotesize]{subfig}
\usepackage{textcomp}
\usepackage{stfloats}
\usepackage{url}
\usepackage{verbatim}
\usepackage{graphicx}
\usepackage{cite}
\usepackage{booktabs}

\usepackage[capitalize]{cleveref}
\crefname{section}{Sec.}{Secs.}
\Crefname{section}{Section}{Sections}
\Crefname{table}{Table}{Tables}
\crefname{table}{Tab.}{Tabs.}

\newcommand{\mten}{\mathbf}

\newcommand{\mset}{\mathbb}

\begin{document}

\title{Transformer-based stereo-aware 3D object detection \\ from binocular images}

\author{
	Hanqing Sun, %
	Yanwei Pang,~\IEEEmembership{Senior Member,~IEEE}, %
	Jiale Cao,~\IEEEmembership{Member,~IEEE}, %
	\\%
	Jin Xie, %
	and %
	Xuelong Li,~\IEEEmembership{Fellow,~IEEE}%
    \thanks{%
		This work was supported by the National Key Research and Development Program of China (No. 2022ZD0160400), %
            and partly supported by the National Natural Science Foundation of China (No. 62401538, 62271346, 62206031). %
		(Corresponding authors: Yanwei Pang, Jiale Cao) %
	}%
	\thanks{%
		Hanqing Sun is with the Changchun Institute of Optics, Fine Mechanics and Physics, Chinese Academy of Sciences, Changchun, China. (E-mail: SunHanqing@ciomp.ac.cn). %
	}%
	\thanks{%
		Yanwei Pang and Jiale Cao are with the Tianjin Key Laboratory of Brain-Inspired Intelligence Technology, School of Electrical and Information Engineering, Tianjin University, Tianjin, China. (E-mail: \{pyw, connor\}@tju.edu.cn). %
	}%
	\thanks{%
		Jin Xie is with the School of Big Data and Software Engineering, Chongqing University, Chongqing, China. (E-mail: xiejin@cqu.edu.cn). %
	}%
	\thanks{%
		Xuelong Li is with the School of Computer Science and School of Artificial Intelligence, Optics and Electronics (iOPEN), Northwestern Polytechnical University, Xi'an, China. (E-mail: li@nwpu.edu.cn). %
	}%
	\thanks{%
		Manuscript received June 14, 2023; %
		revised Jan. 5, 2024; %
		revised April 27, 2024; %
		revised June 27, 2024; %
		accepted Aug. 26, 2024.
	}%
}

\markboth{IEEE Transactions on Intelligent Transportation Systems,~Vol.~XX, No.~XX, September~2024}%
{Hanqing Sun, Yanwei Pang, \MakeLowercase{\textit{et al.}}: Transformer-based stereo-aware 3D object detection from binocular images}%

\maketitle

\begin{abstract}
    Transformers have shown promising progress in various visual object detection tasks, including monocular 2D/3D detection and surround-view 3D detection. %
	More importantly, the attention mechanism in the Transformer model and the 3D information extraction in binocular stereo are both similarity-based. %
	However, directly applying existing Transformer-based detectors to binocular stereo 3D object detection leads to slow convergence and significant precision drops. %
	We argue that a key cause of that defect is that existing Transformers ignore the binocular-stereo-specific image correspondence information. %
	In this paper, we explore the model design of Transformers in binocular 3D object detection, focusing particularly on extracting and encoding task-specific image correspondence information. %
	To achieve this goal, we present TS3D, a Transformer-based Stereo-aware 3D object detector. %
	In the TS3D, a Disparity-Aware Positional Encoding (DAPE) module is proposed to embed the image correspondence information into stereo features. %
	The correspondence is encoded as normalized sub-pixel-level disparity and is used in conjunction with sinusoidal 2D positional encoding to provide the 3D location information of the scene. %
	To enrich multi-scale stereo features, we propose a Stereo Preserving Feature Pyramid Network (SPFPN). %
	The SPFPN is designed to preserve the correspondence information while fusing intra-scale and aggregating cross-scale stereo features. %
	Our proposed TS3D achieves a 41.29\% Moderate Car detection average precision on the KITTI test set and takes 88 ms to detect objects from each binocular image pair.
	It is competitive with advanced counterparts in terms of both precision and inference speed. %
\end{abstract}


\section{Introduction}
\label{sec:ts3d:intro}

\IEEEPARstart{S}{tereo} vision systems can perceive 3D scene for autonomous vehicles and intelligent transportation~\cite{Arnold_Survey3DObject_2019}.
In binocular stereo based 3D perception, image correspondence information plays an essential role~\cite{Zeng_DeepProgressiveFusion_2022,Dinh_FeatureEngineeringDeep_2022,yi_pyramid_2020}. %
Image correspondence information, in the form of disparity or depth supervision, is widely used in convolutional neural network (CNN) based binocular 3D object detectors to prevent the detector from converging to a monocular local minimum~\cite{Liu_YOLOStereo3DStepBack_2021,Chen_DSGNDeepStereo_2020}. %

Vision Transformers~\cite{Dosovitskiy_ImageWorth16x16_2021}, which are equipped with similarity-based attention mechanisms and long-range modeling ability, seem compatible with binocular detection at first glance.
On the one hand, the attention mechanism in Transformers and the critical 3D information extraction in binocular stereo are both similarity-based. %
On the other hand, the Transformer's inherent ability of long-range modeling suites for capturing the wide range of disparities (\textit{e.g.}, a range of $(0, 192]$ in KITTI~\cite{geiger_are_2012}).
In addition, Transformers have gained promising progress in 2D object detection~\cite{Carion_EndtoendObjectDetection_2020,Guo_DenseTrafficDetection_2022,Zhu_MultiModalFeaturePyramid_2023}, surround-view 3D object detection~\cite{Wang_DETR3D3DObject_2021,Liu_PETRPositionEmbedding_2022,Li_BEVFormerLearningBird_2022}, and depth estimation~\cite{Li_RevisitingStereoDepth_2021}. %
Therefore, we focus on adopting the Transformer model to binocular 3D detection in this paper, which can hopefully benefit future studies by enabling binocular detection to interact with Transformer-based language models.
However, for binocular 3D object detection in driving scenes, to the best of our knowledge, there have not been any public Transformer-based detectors in the literature. %

\begin{figure}[!t]
	\centering%
        \subfloat[Curves of Moderate validation AP of DETR3D-Binocular and TS3D\label{fig:ts3d:training:curve}]{%
		\includegraphics[width=0.9\linewidth]{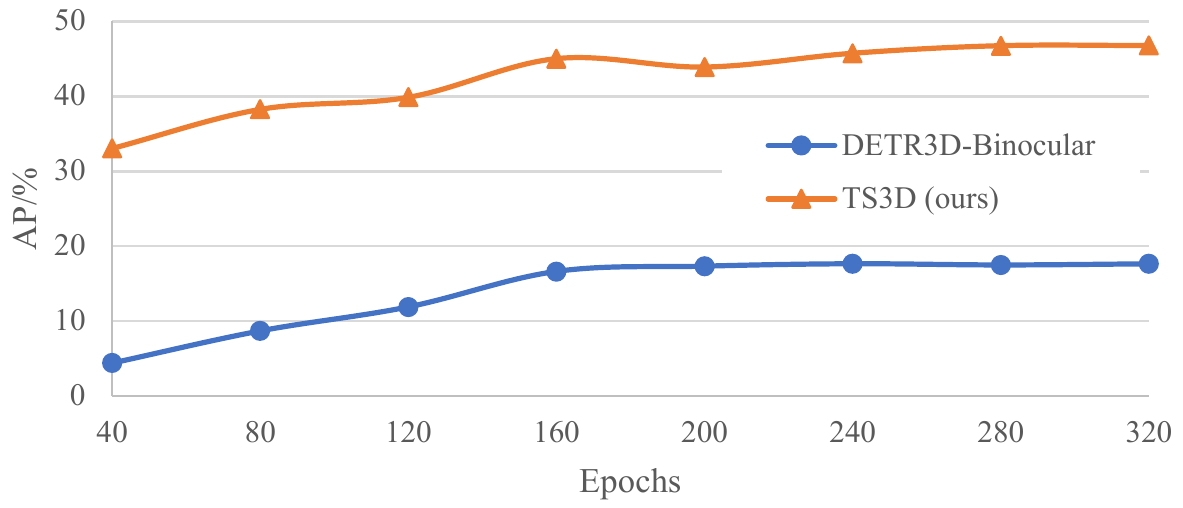}%
	}%
        \\%
        \subfloat[Existing Transformer-based surround-view detectors do not perform well on binocular detection due to the lack of correspondence information. %
		\label{fig:ts3d:training:tab}]{%
		\footnotesize{%
		\begin{tabular}{@{}ccccc@{}}%
			\toprule%
			Method                                           & Epochs & \textit{Mod.}  & Easy           & Hard           \\ \midrule%
			DETR3D~\cite{Wang_DETR3D3DObject_2021}-Binocular & 320    & 17.61          & 29.95          & 13.69          \\%
			PETR~\cite{Liu_PETRPositionEmbedding_2022}-Binocular & 320    & 15.48          & 20.03          & 14.24          \\%
			BEVFormer~\cite{Li_BEVFormerLearningBird_2022}-Binocular & 320    & 23.05          & 30.21          & 21.56          \\%
			TS3D (ours)                                      & 320    & \textbf{46.75} & \textbf{70.50} & \textbf{35.48} \\ \bottomrule%
		\end{tabular}%
		}%
	}%
	\caption{%
		We adapt Transformer-based surround-view 3D object detectors DETR3D~\cite{Wang_DETR3D3DObject_2021}, PETR~\cite{Liu_PETRPositionEmbedding_2022}, and BEVFormer~\cite{Li_BEVFormerLearningBird_2022} to binocular detection. %
        The above Transformer-based detectors are all trained on the KITTI training subset for 320 epochs. %
        The Moderate validation AP of DETR3D-Binocular during training is plotted in (a) and the 3D detection APs are listed in (b). %
        Existing Transformer detectors converge to a poor 3D detection AP, whereas the Transformer-based TS3D can be trained to a superior performance. %
	}%
	\label{fig:ts3d:training}%
\end{figure}

An intuitive way to introduce Transformer models into binocular 3D object detection is to adopt the above surround-view Transformer-based 3D detectors to the binocular scene (\textit{e.g.}, KITTI~\cite{geiger_are_2012}). %
To achieve that, we train binocular revisions of existing surround-view 3D object detectors, which are named with Binocular suffixes. %
Their validation APs after training 320 epochs are enumerated in \cref{fig:ts3d:training:tab}, and the validation AP curve of DETR3D-Binocular during training is shown in \cref{fig:ts3d:training:curve} as blue circles. %
It can be observed that existing Transformer detectors converge to a poor 3D detection performance, which can not be compared to advanced CNN-based binocular 3D object detectors nor their surround-view detection performances~\cite{Caesar_NuScenesMultimodalDataset_2020}. %

We argue that the convergence problem is caused by the lack of image correspondence information in surround-view 3D object detection Transformers such as DETR3D, as well as in existing monocular-based 3D object detection Transformers. %
It is a fact that monocular images do not involve cross-view correspondence~\cite{Huang_MonoDTRMonocular3D_2022,Sheng_Improving3DObject_2021}; %
and the overlapping area across images in a surround-view system is small~\cite{Caesar_NuScenesMultimodalDataset_2020}. %
Concerning the binocular vision system, however, the overlapping areas between left and right images occupy a large proportion of the pixels. %
The latent image correspondence information in the large overlap areas is crucial for 3D scene perception. %
Therefore, extracting and exploiting the image correspondence information in a Transformer-based model is a task-specific challenge of binocular 3D object detection. %

In order to encode the image correspondence information into Transformers, and to build a Transformer-based binocular 3D object detector, we present TS3D, a Transformer-based Stereo-aware binocular 3D object detector. %
The key novelties of the TS3D lie in its image correspondence information preserving (Stereo Preserving Feature Pyramid Network, SPFPN) and encoding (Disparity-Aware Positional Encoding, DAPE) approaches in Transformer models.
It can be used as a generic technique that enables a Transformer design in binocular 3D detection. %

In detail, the SPFPN is proposed to preserve image correspondence information and extract enriched multi-scale stereo features. %
Besides providing semantic information as done in existing feature pyramid networks~\cite{chang_pyramid_2018,wu_semantic_2019,yi_pyramid_2020}, SPFPN preserves both local information in low-level features and correspondence information in stereo features. %
The resultant stereo feature pyramid is then aggregated by intra-scale and cross-scale fusion while respecting the disparity-wise definition of the stereo features. %

Decoder-wisely, a deformable Transformer decoder~\cite{Zhu_DeformableDETRDeformable_2021} is introduced to further aggregate matching costs and decode multi-scale stereo features. %
In the decoder, the DAPE is designed to explicitly encode the sub-pixel-level correspondence information into the stereo features. %
DAPE composites normalized disparities and sinusoidal 2D positional encoding~\cite{Zhu_DeformableDETRDeformable_2021}, thus simultaneously providing position information in both 2D image space and 3D scene. %
In addition to the DAPE, a Non-Parametric Anchor-Based Object Query (NPAQuery) scheme is introduced. %
Compared with widely-used parametric queries, NPAQuery does not introduce new learnable parameters as query embeddings, instead, it reuses the stereo features as a uniformly distributed query embedding in the 2D image space. %

Experiments on the KITTI dataset show that TS3D is an effective Transformer model for binocular 3D object detection, and it is competitive with existing CNN-based counterparts in both terms of accuracy and efficiency. %
Consequently, our TS3D can encode the image correspondence information in the Transformer model, thus alleviating the above convergence problem. %
The validation AP curve of our TS3D is plotted in \cref{fig:ts3d:training:curve} as orange triangles, and its validation results are listed in \cref{fig:ts3d:training:tab}. %
It is shown that after preserving and encoding the image correspondence information by the proposed SPFPN and DAPE respectively, the performance gap is significantly diminished. %
To the best of our knowledge, the proposed TS3D is the first public Transformer-based binocular 3D object detector for driving scenes in the literature. %

The key contributions of this paper can be summarized as three-fold: %
\begin{enumerate}%
	\item[(1)]%
		We present the Transformer-based Stereo-aware 3D object detector (TS3D), which is the first public Transformer-based binocular detector in the literature and can preserve and encode the image correspondence information in binocular images. %
	\item[(2)]%
		The Disparity-Aware Positional Encoding (DAPE) is designed to explicitly encode the sub-pixel-level image correspondence information into stereo features by reusing the disparity predictions, thus providing the Transformer decoder with the 3D scene information. %
	\item[(3)]%
		A Stereo Preserving Feature Pyramid Network (SPFPN) is proposed, which preserves image correspondence information in both intra-scale and cross-scale feature fusions. %
		It provides multi-scale stereo features that hold enriched detail and correspondence information for the Transformer decoder. %
\end{enumerate}

\section{Related work}
\label{sec:ts3d:related}

Existing binocular 3D object detectors are based on convolutional neural networks (CNN) and are categorized in this section. %
Representative Transformer-based 3D detectors for monocular images, LiDAR point clouds, and surround-view images are briefly introduced, and readers are encouraged to refer to \cite{Lahoud_3DVisionTransformers_2022,Mao_3DObjectDetection_2023} for more thorough reviews on Transformers-based 3D object detection. %
Existing feature pyramid networks are also reviewed, which relates to our SPFPN. %

Existing binocular 3D object detectors can be divided into two categories based on their sources of image correspondence information: detectors with and without LiDAR-based image correspondence supervision. %
The former~\cite{Wang_PseudoLiDARVisualDepth_2019,Chen_DSGNDeepStereo_2020,Peng_IDA3DInstancedepthaware3D_2020}, shortened as \emph{Stereo-with-LiDAR}, generates ground-truth image correspondence information based on projected LiDAR point clouds, resulting in accurate yet sparse supervision; %
whereas the latter~\cite{Sun_DispRCNNStereo_2020,Sun_SemanticawareSelfsupervisedDepth_2023,Gao_RealTimeStereo3D_2023}, shortened as \emph{Stereo-without-LiDAR}, does not rely on LiDAR point clouds to generate ground-truth image correspondence information. %
However, the pseudo ground-truths obtained by external stereo matching networks can provide inaccurate supervision in those methods~\cite{Liu_YOLOStereo3DStepBack_2021,Konigshof_Realtime3DObject_2019}. %
Some detectors in this category do not use explicit image correspondence supervision~\cite{Li_StereoRCNNBased_2019,Li_RTS3DRealtimeStereo_2021}. %
Although Stereo-with-LiDAR detectors achieve higher detection precision, developing Stereo-without-LiDAR detectors is still essential as LiDAR devices may not always be affordable or available, such as in stereo endoscopes and traffic cameras. %

Stereo-with-LiDAR 3D object detectors are trained with LiDAR-based image correspondence information. %
Pseudo-LiDAR~\cite{Wang_PseudoLiDARVisualDepth_2019,You_PseudoLiDARAccurateDepth_2019,Qian_EndtoendPseudoLiDARImagebased_2020} formulates the task as a cascade of disparity estimation and LiDAR-based 3D detection, therefore, the image correspondence information is explicitly used in the former step. %
OCStereo~\cite{Pon_ObjectcentricStereoMatching_2020} and IDA-3D~\cite{Peng_IDA3DInstancedepthaware3D_2020} exploit the image correspondence information in an instance-aware manner, that is, disparities in foreground pixels are seen more importantly. %
ZoomNet~\cite{Xu_ZoomNetPartawareAdaptive_2020} also generates instance-wise image correspondence, however, it adopts external 3D models to densify the instance-wise correspondence. %
DSGN~\cite{Chen_DSGNDeepStereo_2020}, PLUMENet~\cite{Wang_PLUMENetEfficient3D_2021}, and CDN~\cite{Garg_WassersteinDistancesStereo_2020} use the image correspondence information as auxiliary supervision to the detection supervision. %
LIGA-Stereo~\cite{Guo_LIGAStereoLearningLiDAR_2021} distillates 3D voxel features extracted from the LiDAR-based correspondence information in a knowledge-distillation~\cite{Hinton_DistillingKnowledgeNeural_2014} manner. %

Stereo-without-LiDAR detectors are trained without LiDAR-based image correspondence information. %
Some detectors in this category use external stereo matching algorithms to generate coarse image correspondence information. %
YOLOStereo3D~\cite{Liu_YOLOStereo3DStepBack_2021} generates coarse disparity maps~\cite{Bradski_OpenCVLibrary_2000} and uses them as a multi-task supervision in parallel with the detection supervision. %
Disp R-CNN~\cite{Sun_DispRCNNStereo_2020} generates instance-wise disparities~\cite{He_MaskRCNN_2017} and utilizes external 3D models to densify such information as done in ZoomNet. %
RT3DStereo~\cite{Konigshof_Realtime3DObject_2019} uses the estimated disparity map as an intermediate image correspondence supervision. %
Some detectors use image correspondence information in an implicit manner or ignore the correspondence. %
Stereo R-CNN~\cite{Li_StereoRCNNBased_2019} and TLNet~\cite{Qin_TriangulationLearningNetwork_2019} only use the image correspondence information from the 3D object annotations. %
RTS3D~\cite{Li_RTS3DRealtimeStereo_2021} implicitly utilizes the correspondence by representing 3D scenes in a 4D feature-consistent embedding space. %

The Transformer-based 2D object detector DETR~\cite{Carion_EndtoendObjectDetection_2020} re-formulates the 2D object detection as a collection prediction task and removes the anchor generation and Non-Maximum Suppression (NMS) processes, resulting in an end-to-end detector. %
DeformableDETR~\cite{Zhu_DeformableDETRDeformable_2021} introduces a deformable attention module, which merely applies the multi-head attention mechanism at several learnable key locations around the reference point. %
It significantly reduces the computational cost of the DETR decoder and converges faster. %
The Transformer-based unary 3D object detector MonoDTR~\cite{Huang_MonoDTRMonocular3D_2022} utilizes auxiliary supervision to implicitly learn depth-aware features and a depth-aware Transformer to mine context and depth information from global features. %
MonoDETR~\cite{Zhang_MonoDETRDepthguidedTransformer_2022} takes 3D object candidates as queries and introduces an attention-based depth encoder to encode depth embeddings. %
A depth cross-attention module is proposed to decode inter-query and query-scene interactions. %
The Transformer-based surround-view 3D object detector DETR3D~\cite{Wang_DETR3D3DObject_2021} uses a CNN to extract unary features of each view and a DETR decoder to predict objects. %
PETR~\cite{Liu_PETRPositionEmbedding_2022} extends DETR3D by generating a 3D coordinate grid in the camera frustum space and fusing the 3D coordinates with the 2D features. %
BEVFormer~\cite{Li_BEVFormerLearningBird_2022} generates object queries on the bird's eye view (BEV) and uses both spatial and temporal information. %
A spatial cross-attention module is proposed to extract BEV features from each view and a temporal self-attention module is designed to fuse historical BEV features. %
In addition to the obvious differences in input modalities comparing our TS3D with the above Transformer-based detectors, the image correspondence information is preserved and encoded into enriched stereo feature pyramids by two novel components, namely SPFPN and DAPE. %

Feature Pyramid Network (FPN)~\cite{lin_feature_2017} exploits the multi-level nature of CNNs to generate multi-scale features, so that the features of each level focus more on objects of the specific scale. %
FPN improves the detection performance of multi-scale 2D object detection. %
RetinaNet~\cite{Lin_FocalLossDense_2017} makes use of more levels of the features than that used in FPN, resulting in higher detection precision at a lower computational cost. %
Original FPN fuses features from lower-resolution to higher-resolution (\textit{i.e.}, top-down), and PAFPN~\cite{liu_path_2018} adds an aggregation network from higher-resolution to lower-resolution features (\textit{i.e.}, bottom-up). %
ASFF~\cite{Liu_LearningSpatialFusion_2019} adaptively fuses multi-scale features and spatially filters information that conflicts across scales. %
BiFPN~\cite{tan_efficientdet_2020} designs a simple and efficient bidirectional FPN, in which a basic structure consisting of top-down and bottom-up paths is repeated to capture enriched multi-scale features. %
RFP~\cite{Qiao_DetectoRSDetectingObjects_2021} recursively reuses the FPN structure and introduces the Atrous Spatial Pyramid Pooling (ASPP) module as a connection module between two FPNs, which improves the performance of 2D object detection. %
The above feature pyramids are carefully designed for unary 2D object detection and do not consider the image correspondence nature of binocular images. %
As a result, they can be applied to generate unary feature pyramids in binocular 3D object detection, nevertheless, directly applying them in stereo features will destroy the latent image correspondence information and affect the accuracy of 3D object detection. %

\section{Methodology}
\label{sec:ts3d:method}

We elaborate on our TS3D in this section. %
The overall architecture is introduced in \cref{sec:ts3d:arch}. %
The proposed SPFPN (\cref{sec:ts3d:stereofpn}) and Transformer decoder (\cref{sec:ts3d:decoder}) result in an encoder-decoder pipeline. %
The DAPE is detailed in \cref{sec:ts3d:decoder:dape} as a positional encoding component in the decoder. %

\subsection{Overall architecture of TS3D}
\label{sec:ts3d:arch}

\begin{figure*}[!t]
	\centering%
	\includegraphics[width=\linewidth]{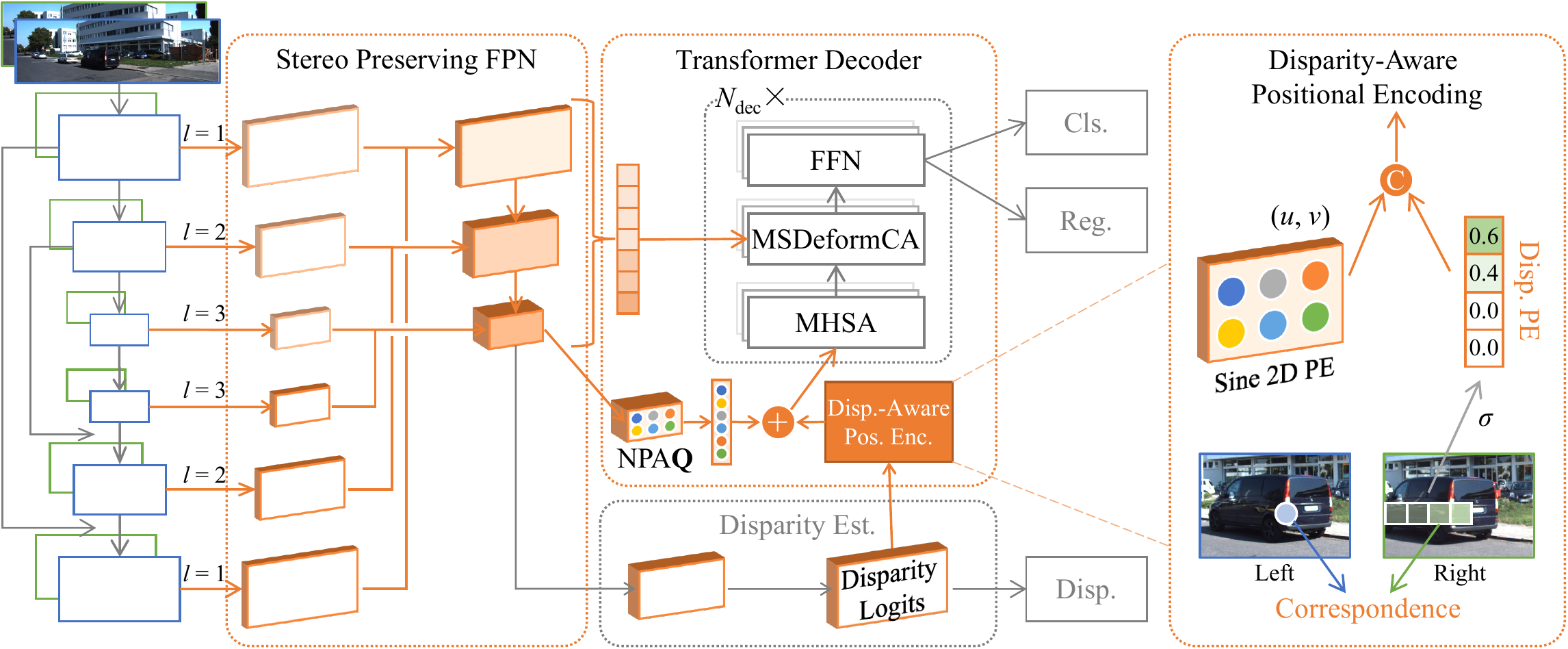}%
	\caption{%
		The overall architecture of the Transformer-based Stereo-aware 3D object detector (TS3D). %
		Sequentially, TS3D takes binocular images as inputs (blue boxes in the figures denote left view, green denotes right), extracts unary features, extracts stereo features using SPFPN (Stereo Preserving Feature Pyramid Network), estimates disparities, decodes object features using a multi-scale deformable DETR decoder~\cite{Zhu_DeformableDETRDeformable_2021}, and regresses and classifies 3D objects. %
		The DAPE (Disparity-Aware Positional Encoding) elaborated on the right is used to explicitly encode image correspondence information for detection. %
	}%
	\label{fig:ts3d:arch}%
\end{figure*}

The overall architecture of TS3D is illustrated in \cref{fig:ts3d:arch}. %
Sequentially, TS3D takes binocular images as inputs (blue boxes in the figures denote left view, green denotes right), extracts unary features and stereo features, estimates disparities, decodes object features using a multi-scale deformable DETR decoder~\cite{Zhu_DeformableDETRDeformable_2021}, finally regresses and classifies 3D objects. %

Two weight-sharing ResNets~\cite{he_deep_2016} $+$ FPN~\cite{lin_feature_2017} are used to \textbf{extract multi-scale unary features} from left and right input images respectively. %
We choose those simple yet effective models as the backbone to preserve detailed information for better image correspondence.
\cref{fig:ts3d:arch} is a schematic diagram of a three-level multi-scale network. For simplicity and clarity, the left and right unary features are represented in a stacked manner (\textit{i.e.}, the stacked blue and green boxes in \cref{fig:ts3d:arch}).
The upper half is the primary unary feature pyramid obtained by the ResNet; the lower half is the enhanced unary feature pyramid obtained by the FPN. %
The index of each feature level in both pyramids is denoted as $l \in \left\{1, 2, 3\right\}$. %

The above unary features are then fed into \textbf{Stereo Preserving Feature Pyramid Network} (SPFPN). %
Since Transformer-based multi-scale stereo matching is time and memory consuming~\cite{Li_RevisitingStereoDepth_2021}, we utilize multi-scale correlation-based cost volumes~\cite{mayer_large_2016} for matching cost computation.
Both primary and enhanced unary feature pyramids are exploited to construct correlation-based cost volumes, resulting in primary and enhanced cost volume pyramids (shown as orange boxes in the second column of \cref{fig:ts3d:arch}). %
Constructed cost volume pyramids are then referred to as stereo feature pyramids in the SPFPN. %
Subsequently, primary and enhanced stereo features from each level of the two pyramids are merged, in which the correspondence information is kept intact. %
The resultant stereo-preserving feature pyramid is then cross-scale aggregated in a bottom-up manner. %
Finally, enriched and aggregated multi-scale stereo features are obtained for subsequent disparity estimation and Transformer decoder. %

A simple \textbf{disparity estimation} head (lower right corner of \cref{fig:ts3d:arch}) is introduced to predict disparity maps from the lowest-resolution stereo feature in the above feature pyramid. %
Stereo features are convolved and upsampled to produce higher-resolution disparity maps. %
The disparity estimation is supervised by pseudo ground-truth generated by the BlockMatching algorithm~\cite{Bradski_OpenCVLibrary_2000}, making the resultant TS3D a Stereo-without-LiDAR 3D object detector. %

The parametric attention module is used for further cost aggregation and multi-scale detection feature extraction.
The \textbf{Transformer decoder} consists of Multi-Head Self-Attention layers~\cite{Carion_EndtoendObjectDetection_2020} (shortened as MHSA in \cref{fig:ts3d:arch}), Multi-Scale Deformable Cross-Attention layers~\cite{Zhu_DeformableDETRDeformable_2021} (MSDeformCA), and Feed-Forward Networks (FFNs). %
The last feature tensor for disparity regression is termed disparity logits and is used as the input of our DAPE. %
The generated positional encoding of the 3D scene is summed with the query of each layer in the Transformer decoder. %
The lowest-resolution features in the above feature pyramid are convolved to generate embeddings of the NPAQuery. %
In addition to the DAPE and NPAQuery, the multi-scale stereo features extracted by SPFPN are exploited as keys and values of the MSDeformCA layers. %
The above processes including MHSA, MSDeformCA, and FFN are collectively referred to as a decoder layer. %
The Transformer decoder is a cascade of $N_\text{dec}$ decoding layers, in which the object queries are refined step-by-step. %
During training, 3D objects are classified and regressed under ground-truth supervision after each decoder layer; %
During inference, only predictions from the last layer are taken as the final 3D object detection results. %

The \textbf{detection head}~\cite{Liu_YOLOStereo3DStepBack_2021} consists of two sub-heads of classification and regression, both of which are predicted at $1/16$ resolution. %
For the classification head, $K+1$ scores are predicted for $K$-class objects, where the extra class denotes the background class. %
Offsets \textit{w.r.t.} anchor boxes are regressed by the regression head. %
The offset is defined as a 13-dimensional vector: the projected 2D bounding box $(u_\text{2D}, v_\text{2D}, w_\text{2D}, h_\text{2D})$, projected 3D object center $(u_\text{3D}, v_\text{3D})$, object distance $z$, object 3D size $(w, h, l)$, and object orientation $(\sin {2 \alpha}, \cos {2 \alpha}, c_\alpha)$. %

\subsection{Stereo Preserving Feature Pyramid Network}
\label{sec:ts3d:stereofpn}

\begin{figure*}[!t]
	\centering%
        \begin{minipage}[b]{0.35\linewidth}%
        \subfloat[FPN\label{fig:ts3d:stereofpn:fpn}]{%
            \includegraphics[height=2.9 cm]{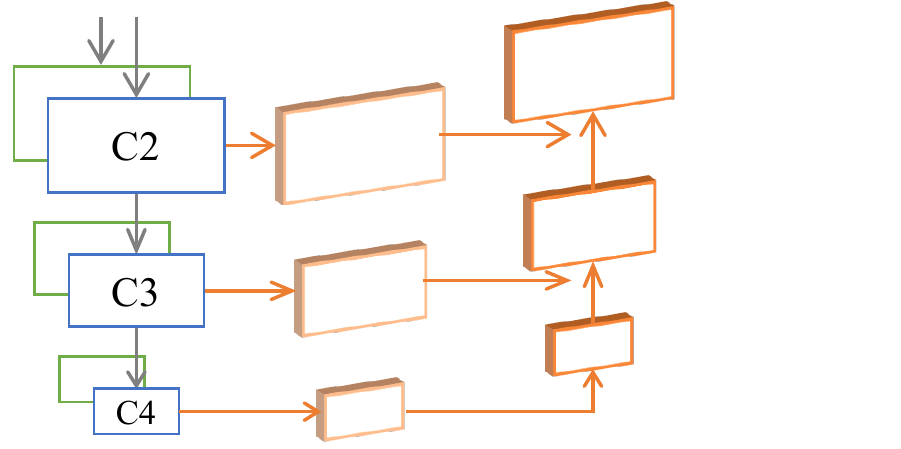}%
        }%
        \\%
        \subfloat[BiFPN\label{fig:ts3d:stereofpn:bifpn}]{%
            \includegraphics[height=2.9 cm]{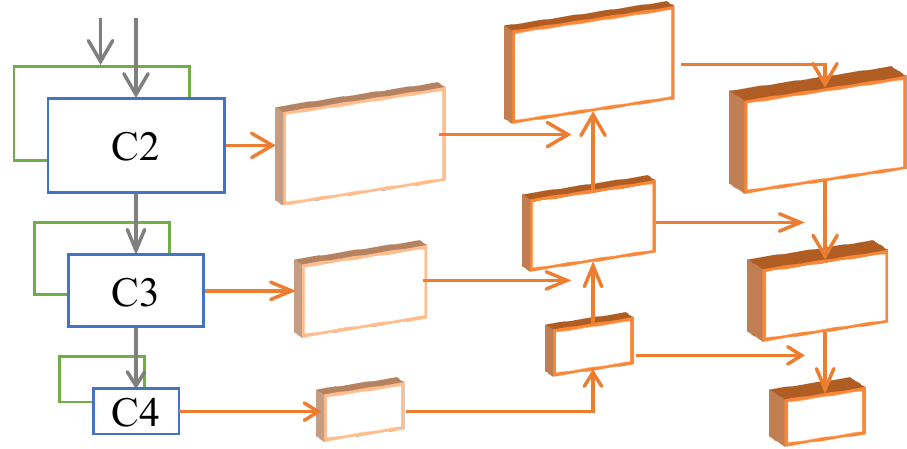}%
        }%
        \end{minipage}%
        \medskip%
        \begin{minipage}[b]{0.64\linewidth}%
        \subfloat[Our SPFPN\label{fig:ts3d:stereofpn:stereofpn}]{%
            \includegraphics[height=6.9 cm]{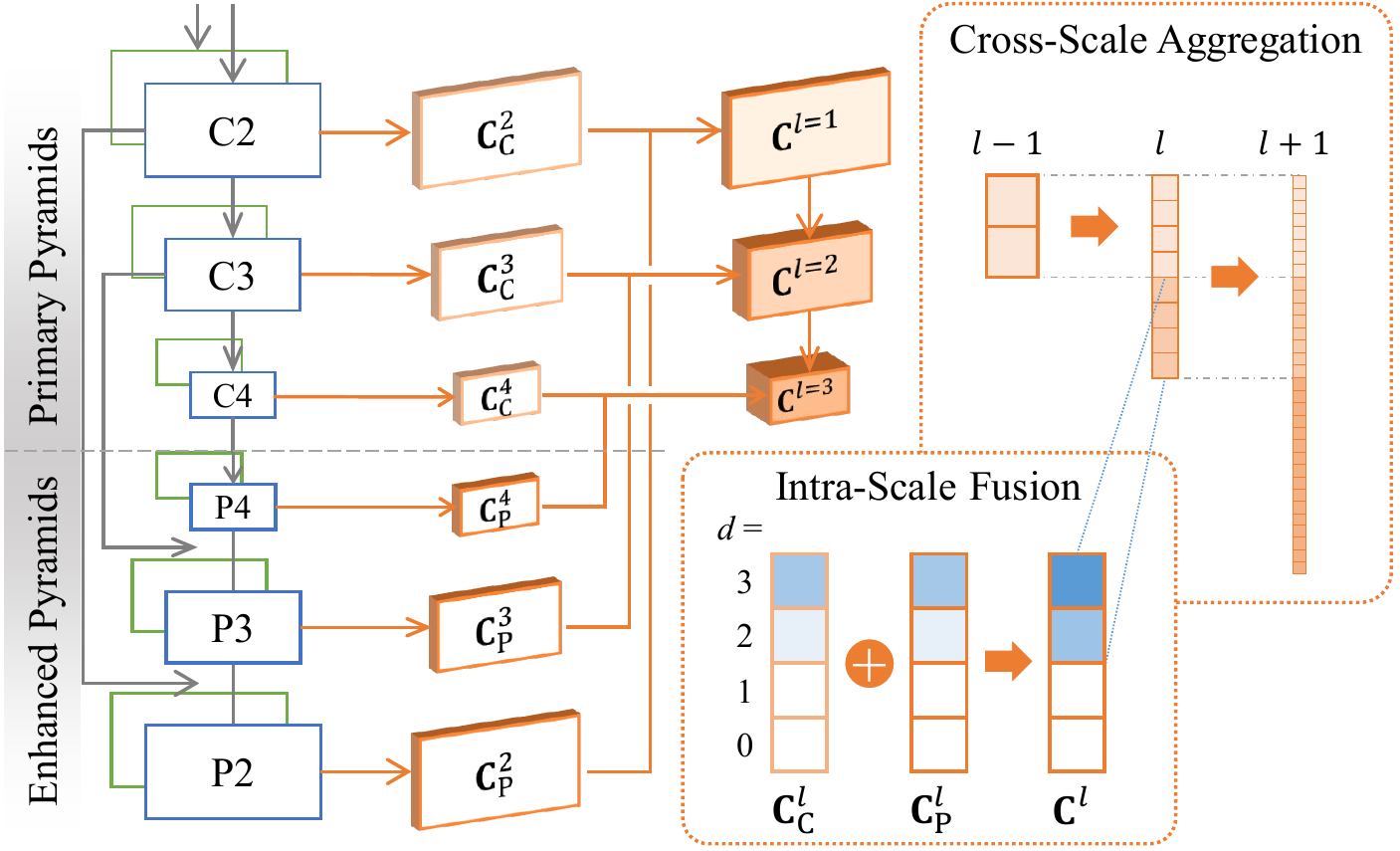}%
        }%
        \end{minipage}%
	\caption{%
		Comparing (a) FPN~\protect\cite{lin_feature_2017} and (b) BiFPN~\protect\cite{tan_efficientdet_2020} with the proposed (c) Stereo Preserving Feature Pyramid Network (SPFPN). %
		FPN consists of a top-down path and BiFPN introduces an additional bottom-up path. %
		Our SPFPN utilizes the FPN to extract multi-scale unary features, and a six-level three-scale cost volume pyramid is constructed from the unary features. %
		Intra-Scale Fusion is performed where disparity dimensions are of identical definition, thus the stereo features are summed accordingly; %
        Cross-Scale Aggregation is performed where disparity dimensions are of different definitions, thus the stereo features are expended and concatenated with the lower-resolution feature. %
		The image correspondence information is therefore preserved. %
	}%
	\label{fig:ts3d:stereofpn}%
\end{figure*}

In order to provide the Transformer decoder with more discriminative stereo features and to embed rich detail and correspondence information into the stereo features, we propose a Stereo Preserving Feature Pyramid Network (SPFPN). %

The feature pyramid network can extract multi-scale features in a neural network model, which is suitable for addressing the scale variation problem in object detection tasks. %
In binocular 3D object detection, on the one hand, unary features suffer from scale variation as in 2D object detection, making the feature pyramid network one of the crucial designs. %
On the other hand, stereo features contain image correspondence information and have clear physical meanings. %
Directly using existing feature pyramid networks for 2D object detection will destroy the 3D information, since those ignore the intra-scale and cross-scale relationships of stereo features. %
Existing designs construct feature pyramids on either unary or stereo features: %
the former cannot perform information interaction across scales of stereo features; %
the latter suffers from the low quality of the initial cost volumes. %

To address those problems, we propose SPFPN, which consists of three parts (three columns of features as shown in \cref{fig:ts3d:stereofpn:stereofpn}). %
The first column presents the unary feature pyramid. %
As described in \cref{sec:ts3d:arch}, the upper part is the primary pyramid (C2--C4) and the lower part is the enhanced pyramid (P2--P4). %
Six levels of paired unary features with three levels of resolutions are used to construct a six-level correlation-based cost volume pyramid~\cite{mayer_large_2016,Sun_SemanticawareSelfsupervisedDepth_2023} (orange rectangles in the second column in the figure). %
The upper three levels in the figure are obtained from primary unary features, which contain rich detailed information; %
the lower three levels are from enhanced unary features, which contain more semantic information. %
The primary and enhanced stereo features are fused according to their resolutions. %
Because the channel dimensions of identical resolution stereo features are of identical physical meanings, those two feature tensors are directly summed. %
The resulting multi-scale stereo features contain rich details and correspondence information. %
In order to get better object embeddings, the higher-resolution stereo features are step-by-step fused with lower-resolution features~\cite{Liu_YOLOStereo3DStepBack_2021} (the downward arrows in the third column of \cref{fig:ts3d:stereofpn:stereofpn}). %
Each scale focuses on objects within the range of the receptive field~\cite{lin_feature_2017}, therefore, the cross-scale fusion in this paper is accomplished by channel expansion and concatenation. %
Finally, the lowest-resolution stereo features with the largest number of feature channels are obtained and reused in NPAQuery. %

Then in primary unary features, the left unary feature $\mten{x}_{\text{L,C}}^{l + 1}$ and the right unary feature $\mten{x}_{\text{R,C}}^{l + 1}$ construct primary 3D cost volumes by%
\begin{equation}
	\mten{C}^{l + 1}_{\text{C}} = \operatorname{CV}_{\text{3D}} \left( \mten{x}_{\text{L,C}}^{l + 1}, \mten{x}_{\text{R,C}}^{l + 1} \right)%
	, %
\end{equation}
where $\operatorname{CV}_{\text{3D}}(\cdot,\cdot)$ correlation based cost volume~\cite{mayer_large_2016}, $\mten{x}_{\text{L,C}}^{l + 1}, \mten{x}_{\text{R,C}}^{l + 1} \in \mset{R}^{\frac{W}{2^{l + 1}} \times \frac{H}{2^{l + 1}} \times C}$, $\mten{C}^{l + 1}_{\text{C}} \in \mset{R}^{\frac{W}{2^{l + 1}} \times \frac{H}{2^{l + 1}} \times \frac{D}{2^{l + 1}}}$, $D$ denotes the maximum disparity value. %
Similarly, the enhanced unary features, $\mten{x}_{\text{L/R,P}}^{l + 1}$ construct enhanced 3D cost volumes. %
It can be observed that $\mten{C}^{l + 1}_{\text{C}}$ and $\mten{C}^{l + 1}_{\text{P}}$ are of identical definition along the disparity dimension so that summation will not change the physical meaning of that dimension. %
Therefore, the initial stereo feature of the $l$-th scale in the third column of \cref{fig:ts3d:stereofpn:stereofpn} is%
\begin{equation}
	\mten{C}^{l}_{\text{init}} = \mten{C}^{l + 1}_{\text{C}} + \mten{C}^{l + 1}_{\text{P}}%
	, %
\end{equation}
$\mten{C}^{l}_{\text{init}} \in \mset{R}^{\frac{W}{2^{l + 1}} \times \frac{H}{2^{l + 1}} \times \frac{D}{2^{l + 1}}}$ is kept. %
However, definitions of the disparity dimension across scales are not identical. %
In order not to destroy the physical meaning of each scale, the cross-scale fusion can be expressed recursively as%
\begin{equation}
	\begin{aligned}%
	\mten{C}^{l} := &\operatorname{Concat} \left[ \mten{C}^{l}, \operatorname{Conv}_{3 \times 3, \text{s.} 2} \left( \mten{C}^{l - 1} \right) \right]%
	, \\%
	\mten{C}^{1} = &\mten{C}^{1}_{\text{init}}%
	, %
	\end{aligned}%
\end{equation}
where $\operatorname{Conv}_{3 \times 3, \text{s.} 2} (\cdot)$ denotes a $3 \times 3$ convolutional layer with a stride of 2. %

Additionally, \cref{fig:ts3d:stereofpn:fpn,fig:ts3d:stereofpn:bifpn} respectively showcase using FPN~\cite{lin_feature_2017} and BiFPN~\cite{tan_efficientdet_2020} for extracting multi-scale stereo features. %
It can be seen from \cref{fig:ts3d:stereofpn:fpn} that the feature fusion direction of FPN is top-down, \textit{i.e.}, using high-level features to supplement semantic information for low-level. %
However, the lowest-resolution stereo features play a vital role in NPAQuery, so we choose to use higher-resolution features to supplement the detailed information of lower-resolution features. %
In this way, it tends to cause a lack of semantic information on high-resolution features. %
Therefore, an hourglass-shaped SPFPN is designed, where the bottom high-resolution features are enhanced by rich semantic information. %
The BiFPN (\cref{fig:ts3d:stereofpn:bifpn}) introduces top-down and bottom-up multi-scale feature fusion. %
Although it can enrich the final multi-scale stereo features, its cross-scale fusion method will destroy the image correspondence information of different scales. %

\subsection{
Transformer decoder%
}
\label{sec:ts3d:decoder}

In order to decode 3D object attributes from stereo features extracted by the above SPFPN, we design a Transformer-based decoder module. %

DeformableDETR~\cite{Zhu_DeformableDETRDeformable_2021} introduces a Multi-Scale Deformable Cross-Attention (MSDeformCA) layer in the DETR decoder~\cite{Carion_EndtoendObjectDetection_2020}. %
We first briefly review the key processes of MSDeformCA as preliminary knowledge in this section. %
Flattened multi-scale features are used as the key and value of MSDeformCA, and object embeddings extracted by MHSA are used as queries. %
A reference point is first extracted from an object query; %
then, MSDeformCA is applied to regress $n_\text{points}$ offset points around it. %
The Multi-Head Cross-Attention (MHCA) mechanism is only applied to those points, which not only reduces the computational cost and GPU memory footprint, but also speeds up the training convergence. %
Therefore, MSDeformCA can be applied to multi-scale features since its computational cost is controllable. %

Despite the success of MSDeformCA in 2D object detection, directly applying it in binocular 3D object detection still leads to a difficult convergence during training. %
To speed up the convergence of such a model, we encode image correspondence information into stereo features. %
Image correspondence information can ensure that the model makes full use of the stereo features, instead of falling into the local minimum of unary 3D object detection~\cite{Chen_DSGNDeepStereo_2020}. %
Towards that end, a Transformer decoder is designed as shown in \cref{fig:ts3d:arch}. %
The main difference between our decoder and a DeformableDETR decoder is Disparity-Aware Positional Encoding (\cref{sec:ts3d:decoder:dape}) and Non-Parametric Anchor-Based Object Query (\cref{sec:ts3d:decoder:query}). %

As shown in \cref{fig:ts3d:arch}, Disparity-Aware Positional Encoding is aimed to explicitly encode image correspondence information based on disparity logits. %
The generated positional encoding holds the 3D location information in the scene. %
It is fed into the decoder as an addition to the NPAQuery, thus introducing 3D information into the Transformer decoder. %
The multi-scale stereo features (described \cref{sec:ts3d:stereofpn}) are flattened and concatenated (in \cref{fig:ts3d:arch}, a small square represents the feature vector of a pixel in the stereo feature, and varying saturations represent features from different scales). %
The resultant 2D feature matrix is used as the multi-scale key/value in MSDeformCA. %
The downsampling ratios of the feature maps of the three resolutions relative to the input image are 4, 8, and 16, respectively, that is, $\mten{x}^l \in \mset{R}^{\frac{W}{ 2^{l+1}} \times \frac{H}{2^{l+1}} \times C_{\text{dec}}}$, where $l \in \left\{ 1, 2, 3 \right\}$, $C_{\text{dec}}$ denotes the dimension of feature vectors in the decoder. %
Only the lowest-resolution stereo features are reused in the NPAQuery, which maintains the inference speed while accelerating the training convergence. %
For other hyperparameters in MSDeformCA, we use $N_\text{dec}=4$ decoder layers, each layer is of $M=8$ heads, and each MSDeformCA samples $K=4$ offset points. %

\subsubsection{Disparity-Aware Positional Encoding (DAPE)}
\label{sec:ts3d:decoder:dape}

Positional encoding is an important component in vision Transformers~\cite{Dosovitskiy_ImageWorth16x16_2021,Liu_PETRPositionEmbedding_2022,Liu_PETRv2UnifiedFramework_2022}. %
The sinusoidal 2D positional encoding~\cite{Carion_EndtoendObjectDetection_2020} is widely used in existing Transformer-based 2D object detectors. %

However, positional encodings for 2D object detection lack 3D scene information; positional encodings for unary, point cloud, and surround-view 3D object detection ignore the correspondence information between binocular images. %
Therefore, DAPE is proposed to explicitly encode the image correspondence information into stereo features based on disparities, thus enabling the decoder to perceive the 3D information of the scene and objects. %

We reuse the disparity logits in the disparity estimation head. %
The disparity prediction at position $(u, v)$ on the disparity map $\hat{\mten{M}}$ is regressed using SoftArgMax~\cite{chang_pyramid_2018} by the disparity logits $\mten {x}_{\hat{\mten{M}}}$: %
\begin{equation}
	\hat{\mten{M}} \left( u, v \right) = \operatorname{SoftArgMax} \left( \mten{x}_{\hat{\mten{M}}} \left( u, v, : \right) \right)%
	, %
\end{equation}
where $\mten{x}_{\hat{\mten{M}}} \in \mset{R}^{W \times H \times C_\text{disp}}$, and $C_\text{disp} < C_\text{dec}$. %
We proposed to reuse the Softmax normalized version of $\mten{x}_{\hat{\mten{M}}}$ as the probability distribution of disparity at $(u, v)$, thus introducing the correspondence information into the decoder. %
Formally, it can be given as%
\begin{equation}
	\operatorname{{PE}_\text{disp}} \left( \mten{x}_{\hat{\mten{M}}}; u, v, : \right) = \sigma \left( \mten{x}_{\hat{\mten{M}}} \left( u, v, : \right) \right)%
	, %
	\label{eq:dape:softmax}%
\end{equation}
where $\sigma \left( \cdot \right)$ denotes the Softmax normalization. %
The DAPE is a concatenation of the sinusoidal 2D and the disparity-related encoding defined by \cref{eq:dape:softmax}, that is, %
\begin{equation}
	\operatorname{{PE}_\text{DA}} \left( \mten{x}_{\hat{\mten{M}}}; u, v, : \right) = \left[ \operatorname{{PE}_\text{sine}} \left( u, v, : \right), \operatorname{{PE}_\text{disp}} \left( \mten{x}_{\hat{\mten{M}}}; u, v, : \right) \right]%
	, %
	\label{eq:dape}%
\end{equation}
where $\operatorname{{PE}_\text{disp}} \left( \mten{x}_{\hat{\mten{M}}}; u, v, : \right)$ is a $C_\text{disp}$-dimensional vector, $\operatorname{{PE}_\text{sine}} \left( u, v, : \right)$ is re-defined as a $( C_\text{dec}-C_\text{disp} )$-dimensional vector, so $\operatorname{{PE}_\text{DA}} \left( \mten{x}_{\hat{\mten{M}}}; u, v, : \right)$ is a $C_\text{dec}$-dimensional vector, which can be summed directly with the object query. %

DAPE is added to the NPAQuery to provide explicit image correspondence and 3D position information for stereo features: %
\begin{equation}
	\mten{Q} = \mten{x}_q + \operatorname{{PE}_\text{DA}} \left( \mten{x}_{\hat{\mten{M}}} \right)%
	, %
\end{equation}
where $\mten{x}_q \in \mset{R}^{\frac{W}{16} \times \frac{H}{16} \times C_{\text{dec}}}$ is the NPAQuery elaborated in \cref{sec:ts3d:decoder:query}, and $\mten{x}_q$ and $\operatorname{{PE}_\text{DA}} \left( \mten{x}_{\hat{\mten{M}}} \right)$ are reshaped to $\mset{R}^{(\frac{W}{16} \cdot \frac{H}{16}) \times C_{\text{dec}}}$. %

It can be seen from \cref{eq:dape:softmax,eq:dape} that the DAPE not only encodes 2D information in the 2D image space but also explicitly encodes the correspondence information of binocular images. %
It is suitable for locating the projection of 3D object center points and bounding boxes in 2D images; %
The encoded image correspondence information is guided by stereo features, and thus is suitable for regressing depths and sizes of 3D objects in 3D space. %

\subsubsection{Non-Parametric Anchor-Based Object Query (NPAQuery)}
\label{sec:ts3d:decoder:query}

Existing methods assign a best-matching object query to each object. %
When the number of positive samples is small (as that in KITTI), few object queries can be trained in each iteration, resulting in less training efficiency. %
As a consequence, it is difficult for the model to learn useful 3D object features. %
In addition, in order to make object queries cover the entire 3D scene, existing 3D object detectors usually generate dense 3D grids. %
Therefore, the number of object queries becomes large to ensure that it can cover the entire 3D space, which makes the training harder to converge. %

To alleviate that problem while ensuring the object query able to cover the entire 3D space, we reuse the lowest-resolution features in the pyramid (\cref{sec:ts3d:stereofpn}). %
The feature map can cover the entire 2D image space, and with the 3D information provided by the proposed DAPE, it can cover the 3D space. %
The NPAQuery does not depend on learnable parameters as do in \cite{Qian_EndtoendPseudoLiDARImagebased_2020,Li_DNDETRAccelerateDETR_2022}, thus we term it a non-parametric object query. %
Using higher-resolution feature maps can cover the 3D space more densely, but it also significantly increases the computational cost. %
In terms of query quantity and coverage, the lowest-resolution stereo feature is a compromise. %

As mentioned above, the lowest-resolution feature $\mten{x}^{l=3} \in \mset{R}^{\frac{W}{16} \times \frac{H}{16} \times C_{ \mten{x}}^3}$. %
The number of feature channels is convolved to $C_{\text{dec}}$ by a $1 \times 1$ convolution, and the resulting tensor is denoted as $\mten{x}_q \in \mset{R}^{ \frac{W}{16} \times \frac{H}{16} \times C_{\text{dec}}}$, which is used as object embeddings. %
After flattening, the query becomes $\mten{x}_q \in \mset{R}^{ 1440 \times C_{\text{dec}}}$, which is equivalent to 1440 object queries: $N_q = \frac{W}{16} \times \frac{H}{16} = 1440$. %
Although this number is much higher than empirical values in 2D object detection ~\cite{Zhu_DeformableDETRDeformable_2021}, it is comparable to the values in surround-view 3D object detection~\cite{Liu_PETRPositionEmbedding_2022}. %

Because the coordinate correspondence between the NPAQuery and the 2D image space is clear, we therefore directly generate the reference point coordinates according to this correspondence. %
Such design makes it possible to exploit anchor prior~\cite{Liu_YOLOStereo3DStepBack_2021}, which is commonly used in CNN-based detectors. %
During training, targets are assigned \textit{w.r.t.} 2D IoUs (Intersection over Union) between anchor boxes and ground-truth boxes. %
If an IoU is greater than a preset foreground threshold $\tau_{\text{fg}}$, the corresponding anchor is assigned as a matching positive sample; %
On the contrary, if an IoU is smaller than a preset background threshold $ \tau_{\text{bg}}$, the anchor is assigned as a negative sample. %
During training, we set $\tau_\text{fg} = 0.5$ and $\tau_\text{bg} = 0.4$. %

\subsection{Training loss}
\label{sec:ts3d:loss}

We use Focal Loss~\cite{Lin_FocalLossDense_2017} for object classification: %
\begin{equation}
	\mathcal{L}_\text{cls} \left( \hat{p}, p \right) = \left\{%
		\begin{array}{ll}%
			-\alpha \left( 1-\hat{p} \right)^\gamma \log\left( \hat{p} \right), & \text{if } p = 1, \\%
			- \hat{p}^\gamma \log\left( 1-\hat{p} \right), & \text{if } p = 0, %
		\end{array}%
		\right. %
\end{equation}
where $\hat{p}$ is the predicted probability and $p$ denotes the groundtruth label. %
During training, we set balancing weight $\alpha = 20$ and focusing parameter $\gamma = 2$. %
Smoothed L1 Loss is used for bounding-box regression: %
\begin{equation}
	\mathcal{L}_\text{reg} \left( \hat{x}, x \right) = \left\{%
		\begin{array}{ll}%
			0.5 \left( \hat{x} - x \right)^2 / \beta, & \text{if } \left| \hat{x} - x \right| < \beta, \\%
			\left| \hat{x} - x \right| - 0.5 \beta, & \text{otherwise}. %
		\end{array}%
	\right. %
\end{equation}
where $\hat{x}$ is the predicted 3D bounding-box, $x$ denotes the groundtruth bounding-box, and $\beta = 0.04$. %
For object orientation prediction~\cite{Liu_YOLOStereo3DStepBack_2021}, we use Binary Cross Entropy Loss: %
\begin{equation}
	\mathcal{L}_\text{orient} \left( \hat{\alpha}, \alpha \right) = -\left( 1-\hat{\alpha} \right) \log\left( \hat{\alpha} \right)%
	, %
\end{equation}
where $\hat{\alpha}$ is the predicted orientation probability and $\alpha$ denotes the groundtruth label of an orientation bin. %
In the auxiliary disparity head, Stereo Focal Loss~\cite{Zhang_AdaptiveUnimodalCost_2020} is exploited: %
\begin{equation}
	\mathcal{L}_\text{disp} = \sum_{d=0}^{D-1} - P(d)\log \hat{P}(d)%
	, %
\end{equation}
where $\hat{P}(d)$ is the Softmax prediction of $d$-th disparity candidate, $D$ denotes the number of disparity candidates, $P(d)$ is defined as%
\begin{equation}
	P(d) = \frac{\exp\left(-\left| \hat{d} - d \right| / \sigma\right)}{\sum_{d^\prime=0}^{D-1}\exp\left(-\left| \hat{d^\prime} - d^\prime \right| / \sigma\right)}%
	, %
\end{equation}
where temperature $\sigma = 0.5$, $\hat{d}$ is the predicted disparity, and $d$ is the groundtruth disparity. %
Total loss is defined as the summation of the above losses: %
\begin{equation}
	\begin{aligned}%
		\mathcal{L} = \sum_i^{N_\text{dec}} \frac{1}{\left|\mathcal{O}\right|} &\sum \left( \mathcal{L}_{\text{cls};i}%
		+ \mathcal{L}_{\text{reg};i}%
		+ \mathcal{L}_{\text{orient};i} \right) \\%
		+ \frac{1}{HW} &\sum \mathcal{L}_\text{disp}%
		, %
	\end{aligned}%
\end{equation}
where $\mathcal{O}$ denotes the object set in a stereo image pair, $H$ and $W$ respectively denote the height and width of an image, and $N_\text{dec}$ is the number of Transformer decoders. %

\section{Experiments}
\label{sec:ts3d:exp}

\subsection{Experimental setup}
\label{sec:ts3d:exp:setup}

The experiments are conducted on the KITTI dataset~\cite{geiger_are_2012}. %
In the ablation experiments, the data split is consistent with that of \cite{Chen_Multiview3DObject_2017} and the evaluation index is obtained on its validation subset. %
We report all difficulty levels (Easy, Moderate, Hard) of 3D Car detection AP (Average Precision) as benchmarking indicators. %
In the experiments in \cref{sec:ts3d:exp:result}, the entire training set is used during training, and the prediction results are calculated by the KITTI server. %
Detection APs of three difficulty levels of 3D Car and Pedestrian detection tasks are reported on the KITTI test set. %
Among them, Moderate 3D Car detection AP is the main evaluation index in this paper, which is placed as the first column of AP in each table and marked in italics. %

The TS3D model is implemented based on PyTorch 1.10.0~\cite{paszke_pytorch_2019} and Python 3.7.11. %
During training, four NVIDIA Quadro P6000 GPUs are used with CUDA 11.1.74 and CuDNN 8.0.5 installed. %

In the ablation experiments, ResNet-34~\cite{he_deep_2016} is used as the backbone. %
The disparity dimensions of multi-scale cost volumes are set to 24, 48, and 96, respectively, from highest to lowest resolution. %
The experiments on the KITTI test set use ResNet-50 as the backbone, the disparity dimensions are 48, 48, and 96, respectively. %
The rest hyperparameters remain unchanged. %

During training, 8 samples are input on each GPU for each iteration, resulting in an equivalent batchsize of 32 on 4 GPU parallel training. %
The resolution of the input RGB image is uniformly cropped to $W=1280$, $H=288$. %
During the training process, random distortion is applied to the image for data argumentation, including random contrast, hue, saturation, and brightness. %
Training samples are randomly horizontally flipped with a probability of 0.5. %
The learning rate is initially set to $2 \times 10^{-4}$ with weight decay of $1 \times 10^{-4}$. %
We use AdamW~\cite{Loshchilov_DecoupledWeightDecay_2019} algorithm to minimize the loss with the Cosine Annealing learning rate update scheme. %

\subsection{Ablation study}
\label{sec:ts3d:exp:ablation}

Ablation study is performed on the KITTI validation set, and experimental results are shown in \cref{tab:ts3d:exp:ablation}. %
The first row shows the result of directly using the Transformer-based surround-view 3D object detector DETR3D~\cite{Wang_DETR3D3DObject_2021} to train on KITTI. %
The training process is unstable and the detection accuracy is poor. %
The DAPE (second row) is used in conjunction with NPAQuery to enrich the stereo features, and the Moderate AP is greatly improved by 27.98 p.p. (percentage points), and other indicators are improved as well. %
Finally, the SPFPN (third row) is introduced to further introduce the detail and correspondence information into stereo features, which improves the Moderate detection AP by 1.17 p.p. %
As a comparison, we also implement the SPFPN on the CNN-based YOLOStereo3D~\cite{Liu_YOLOStereo3DStepBack_2021} (fourth row), and the results are given in the fifth row. %
Compared with baseline results on the fourth row, using SPFPN gains an improvement in detection AP (1.21 p.p.), proving that SPFPN is also effective in CNNs. %
Thanks to its image correspondence information preserving design, the extracted stereo features are enriched and more discriminative. %

\begin{table}
	\centering%
	\caption{%
		Ablation study on the KITTI validation set. %
	}%
	\label{tab:ts3d:exp:ablation}%
	\begin{tabular}{@{}ccc|ccc@{}}%
        \toprule%
        Transformer & DAPE $+$ NPAQuery & SPFPN      & \textit{Mod.}  & Easy           & Hard           \\ \midrule%
        \checkmark  &            &            & 17.61          & 29.95          & 13.69          \\%
        \checkmark  & \checkmark &            & 45.59          & \textbf{70.98}          & 34.71          \\%
        \checkmark  & \checkmark & \checkmark & \textbf{46.76} & 70.90 & \textbf{35.94} \\ \midrule%
                    &            &            & 44.06          & 68.94          & 33.24          \\%
                    &            & \checkmark & \textbf{45.27} & \textbf{70.69} & \textbf{35.77} \\ \bottomrule%
        \end{tabular}%
\end{table}

\subsubsection{Design of positional encoding}

\begin{table}
	\centering%
	\caption{%
		Comparing our Disparity-Aware Positional Encoding (DAPE) with existing positional encodings. %
	}%
	\label{tab:ts3d:exp:dape}%
	\begin{tabular}{@{}cccc@{}}%
	\toprule%
	Positional Encoding      & \textit{Mod.}  & Easy           & Hard           \\ \midrule%
	w/o PE           & 44.63          & 69.76          & 33.88          \\%
	Sine2D PE        & 45.17          & 70.82          & 34.38          \\%
	One-Hot Disparity PE & 43.54          & 67.18          & 32.95          \\%
	DAPE (ours) & \textbf{45.59} & \textbf{70.98} & \textbf{34.71} \\ \bottomrule%
	\end{tabular}%
\end{table}

The DAPE is compared with the widely-used sinusoidal 2D positional encoding and one-hot disparity encoding. %
The results are listed in \cref{tab:ts3d:exp:dape}. %
Specifically, the sinusoidal 2D positional encoding in the table is a fixed embedded 2D positional encoding. %
The one-hot disparity positional encoding is a simple way to explicitly encode disparity information into stereo features. %
First, use ArgMax to find the candidate disparity on the disparity map, and then set the positional embedding of the disparity value index position to 1, and the other parts set to 0. %
As a consequence, it can introduce pixel-level correspondence information into stereo features. %
No positional encodings are introduced in the first row of \cref{tab:ts3d:exp:dape}, which serves as a comparison baseline. %
The second row shows that using sinusoidal 2D positional encoding can introduce location information and the detection AP is improved (0.54 p.p.), however, it does not introduce image correspondence information. %
The one-hot disparity positional encoding in the third row explicitly introduces the correspondence information. %
Results prove that considering merely the pixel-level correspondence information while ignoring the sub-pixel-level correspondence and 2D location information is harmful to 3D object detection. %
Ignoring this fact and forcing the introduction of discretized disparity coding will bring noise to the stereo feature. %
The DAPE (last row) of this paper considers both 2D location information and sub-pixel-level stereo matching information but does not enforce the discretization of disparity. %
It uses the latent correspondence information in stereo features to guide the disparity-aware positional embedding. %
DAPE is a learnable binocular 3D positional encoding, in contrast to the fixed sinusoidal 2D positional encoding. %
Experiments show that DAPE outperforms the above three schemes and can provide suitable 3D spatial information for binocular 3D object detection. %

\begin{figure*}
	\centering%
	\includegraphics[width=\linewidth]{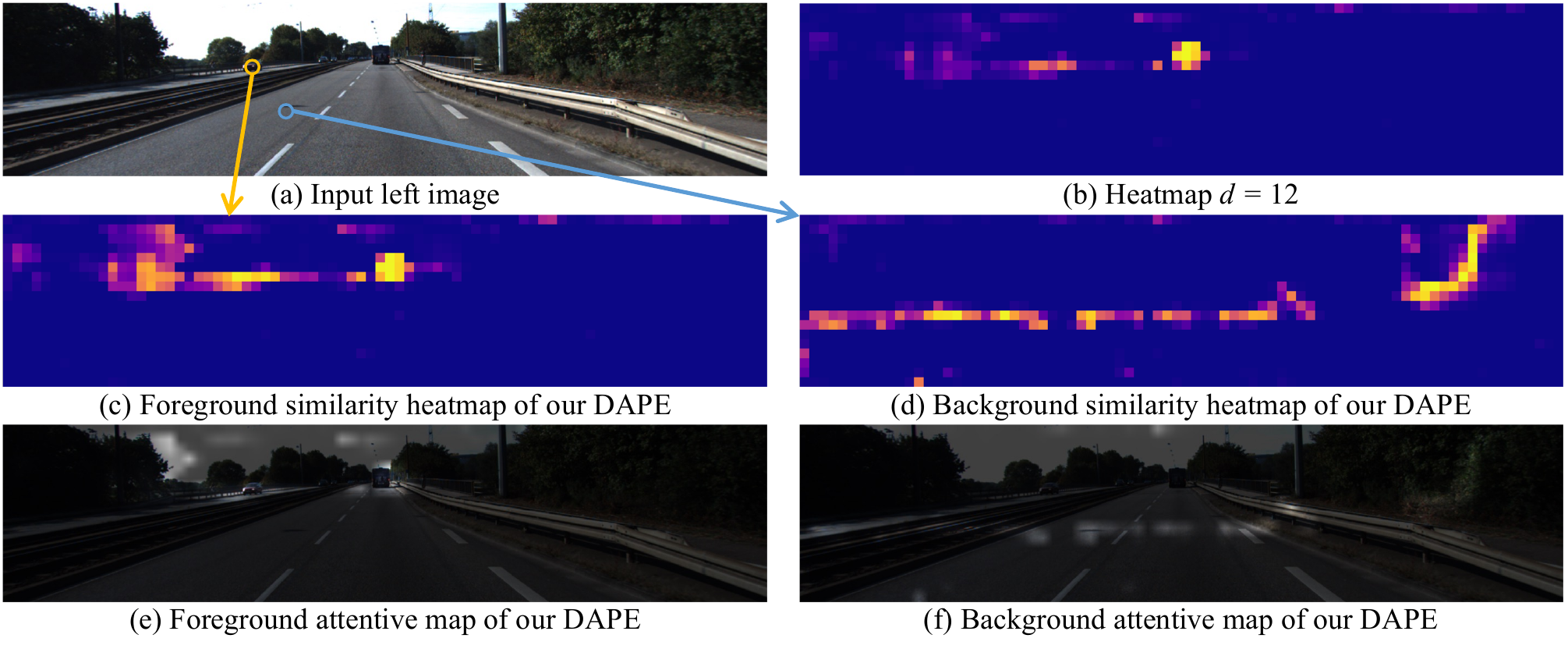}%
	\caption{%
		The characteristics of the proposed DAPE (Disparity-Aware Positional Encoding, see \cref{sec:ts3d:decoder:dape}). %
		(a) Given an input left image, we sample a foreground pixel (yellow circle) and a background pixel (blue circle).
		The disparity of the foreground pixel is approximately $12$, thus the DAPE heatmap at $d = 12$ is visualized as (b).
		For the foreground and background pixels, their positional encodings are respectively dot producted by the encodings at all pixels, and the resultant heatmaps are visualized as (c) and (d), respectively.
		We then respectively mask the input left image with those heatmaps in (e) and (f), demonstrating that DAPE focuses on the areas with similar disparity distributions.
		}%
	\label{fig:ts3d:exp:dape}%
\end{figure*}

Qualitatively, we also illustrate the characteristics of our DAPE in \cref{fig:ts3d:exp:dape}.
Given an input left image, we sample a foreground pixel (yellow circle in \cref{fig:ts3d:exp:dape}a) and a background pixel (blue circle).
The disparity of the foreground pixel is approximately $12$, thus the DAPE heatmap at $d = 12$ is visualized as (b).
For the foreground and background pixels, their positional encodings are respectively dot-producted by the encodings at all pixels to emulate the similarity-based attention mechanism.
The resultant foreground and background positional encoding similarity heatmaps are visualized as (c) and (d), respectively.
To see the attended areas more clearly, we respectively mask the input left image with those heatmaps in (e) and (f).
Results show that our DAPE can encode 3D information into the stereo features and attends well on the foreground objects, since the disparities on the surface of an object are of similar distributions.

\subsubsection{%
	SPFPN compared with existing counterparts%
}

In this subsection, a variety of existing feature pyramid networks are implemented to extract multi-scale stereo features. %
A comparison of those and our SPFPN can be seen in \cref{tab:ts3d:exp:fpn}. %
The first row shows the result of applying FPN~\cite{lin_feature_2017} to extract multi-scale stereo features (detailed in \cref{sec:ts3d:stereofpn}). %
The results show that the direct use of FPN will lead to a significant decline in APs. %
The reason, as mentioned above, is that these methods do not consider the physical meaning of the disparity dimension of stereo features. %
Many FPNs are aimed to enrich features of higher-resolution (\textit{i.e.}, top-down aggregation), which is not compatible with the idea of reusing lowest-resolution features in this paper. %
BiFPN~\cite{tan_efficientdet_2020} (\cref{sec:ts3d:stereofpn}) uses high-level features to supplement the semantic information of low-level features, and then aggregates features to high-level features from the bottom up. %
Therefore, the results in the second row show that the fusion of multi-scale stereo features based on BiFPN can significantly improve the detection AP compared with the above three existing pyramid networks. %
The last row in \cref{tab:ts3d:exp:fpn} shows the detection AP of the model based on our SPFPN. %
Results prove that SPFPN is more suitable for extracting multi-scale stereo features and can retain crucial 3D scene information, providing the Transformer decoder with more discriminative object queries. %

\begin{table}
	\centering%
	\caption{%
		Comparing our Stereo Preserving Feature Pyramid Network (SPFPN) with existing feature pyramids. %
	}%
	\label{tab:ts3d:exp:fpn}%
	\begin{tabular}{@{}ccccc@{}}%
        \toprule%
        Pyramid           & Aggregation          & \textit{Mod.}  & Easy           & Hard           \\ \midrule%
        FPN~\cite{lin_feature_2017}        & Top-down             & 34.49          & 51.92          & 26.61          \\%
        BiFPN~\cite{tan_efficientdet_2020} & Top-down \& bottom-up & 44.19          & 67.37          & 33.72          \\%
        SPFPN (ours)                              & Stereo Preserving \& bottom-up            & \textbf{46.76} & \textbf{70.90} & \textbf{35.94} \\ \bottomrule%
        \end{tabular}%
\end{table}

\subsubsection{The number of decoder layers and the usage of intermediate supervision}

This subsection gradually changes the number of decoding layers in the Transformer decoder, that is, $N_{\text{dec}} \in \left\{2, 4, 6, 8 \right\}$. %
The first row of \cref{tab:ts3d:exp:layer_super} is the detection AP of the baseline model without Transformer decoder (\textit{i.e.}, $N_{\text{dec}} = 0$). %
When the decoder is added without additional supervision (third row), a significant drop in detection AP can be observed. %
Conversely, by adding appropriate supervision (\cref{sec:ts3d:arch}) at each decoding layer, the detection AP can be improved regardless of the number of decoder layers. %
Continuing to increase the number of decoder layers can improve detection AP, however, it requires longer training time. %
We set $N_{\text{dec}} = 4$ in the final model. %

\begin{table}
	\centering%
	\caption{%
		The impact of the number of decoder layers and intermediate supervisions. %
	}%
	\label{tab:ts3d:exp:layer_super}%
	\begin{tabular}{@{}cc|ccc@{}}%
	\toprule%
	\begin{tabular}[x]{@{}c@{}}Decoder\\\#Layer\end{tabular} & \begin{tabular}[x]{@{}c@{}}Intermediate\\Supervision\end{tabular} & \textit{Mod.}  & Easy           & Hard           \\ \midrule%
	$N_{\text{dec}} = 0$ &            & 45.27          & 70.69          & 35.77          \\%
	$N_{\text{dec}} = 2$ & \checkmark & 45.54          & 70.71          & 35.87          \\%
	$N_{\text{dec}} = 4$ &            & 42.79          & 63.68          & 32.68          \\%
	$N_{\text{dec}} = 4$ & \checkmark & 46.76          & \textbf{70.90} & 35.94          \\%
	$N_{\text{dec}} = 6$ & \checkmark & \textbf{46.79} & 70.34          & 35.44          \\%
	$N_{\text{dec}} = 8$ & \checkmark & 46.71          & 70.49          & \textbf{36.01} \\ \bottomrule%
	\end{tabular}%
\end{table}

\subsubsection{%
Discussion on performance drops of existing surround-view Transformer detectors%
}

Broadly speaking, a surround-view scene is of large field-of-view (FOV) and adjacent views are with small overlapping areas;
A classic binocular stereo scene is of relatively smaller FOV and the overlapping areas are large, which enables image correspondence to represent more precise 3D information.
(1) Adapting surround-view detectors to the binocular scene narrows the FOV and introduces large overlaps.
However, the former results in less 3D scene information, and the latter is not fully considered in existing surround-view detectors.
As a consequence, the detection precision significantly drops (shown in \cref{fig:ts3d:training:tab}).
We tackle that problem by introducing image correspondence representations to the Transformer model.
(2) Another reason for the performance drops is insufficient training samples.
Existing Transformer-based multi-view 3D object detectors are trained with 1.4 million object annotations on the nuScenes dataset.
However, the binocular dataset, KITTI, contains merely $1/7$ object annotations of those in nuScenes.
In our TS3D, that problem is alleviated by the NPAQuery.

\subsection{Comparison with existing methods}
\label{sec:ts3d:exp:result}

The performance of our TS3D on the KITTI test set is given in this subsection. %
The evaluation indicators of three difficulty levels (Easy, Moderate, Hard) including 3D Car and Pedestrian detection tasks are displayed in the last row of \cref{tab:ts3d:exp:result}. %

\begin{figure*}[!t]
	\centering%
	\includegraphics[width=\linewidth]{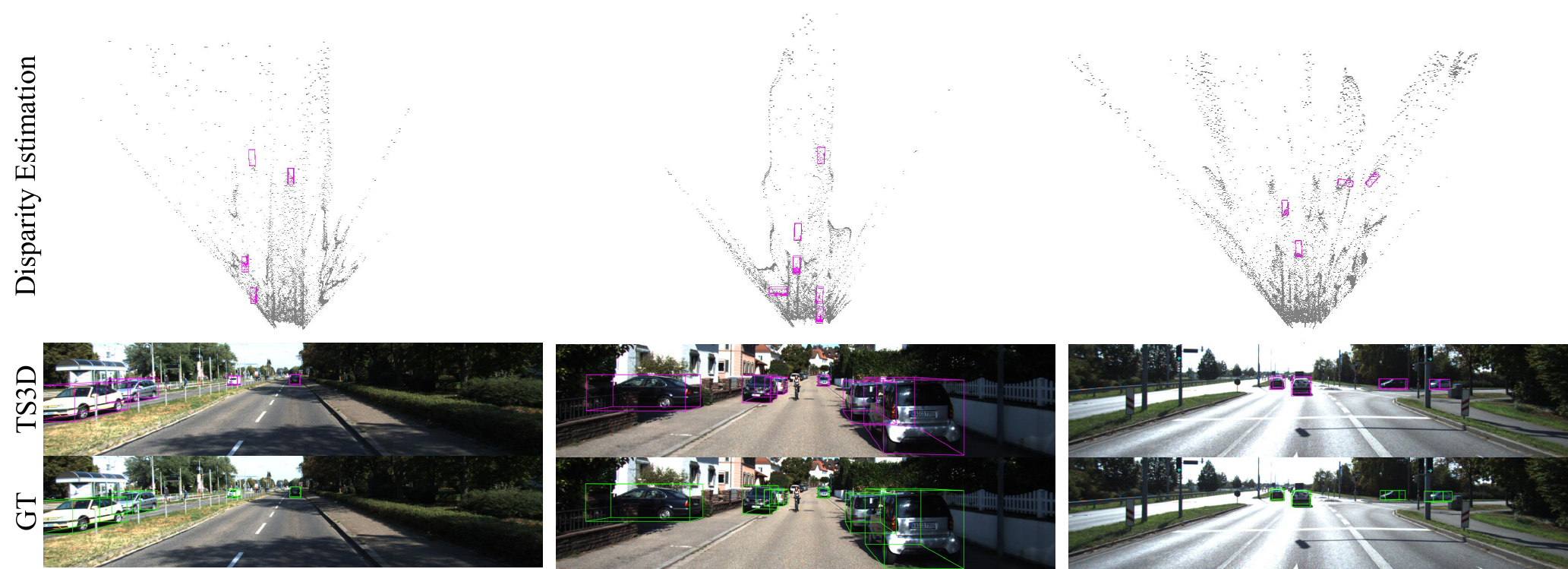}%
	\caption{%
		Visualization of three detection results of our TS3D on the KITTI validation set, one column each. %
        From top to bottom of each sample: inverse-projection of disparity estimation and 3D detection (pink), left image with projected 3D detection (pink), and left image with 3D ground-truth boxes (green). %
	}%
	\label{fig:ts3d:results:success}%
\end{figure*}

\begin{table*}
	\centering%
	\caption{%
		Results on the KITTI test set of our TS3D. %
        It is competitive with advanced CNN-based detectors in terms of both AP and speed. %
	}%
	\label{tab:ts3d:exp:result}%
	\begin{tabular}{rrrr@{}l@{}rrrcl@{}}
		\toprule
		\multicolumn{1}{@{}l}{\textbf{Model - Category}}                 & \multicolumn{3}{c@{}}{3D Car Detection} &  & \multicolumn{3}{@{}c}{3D Pedestrian Detection} & Runtime &                                                       \\ \cmidrule(l){2-4}\cmidrule(r){6-8}
		Method                                                        & \textit{Mod.}    & Easy    & Hard    & \phantom{abcd} & \textit{Mod.}      & Easy       & Hard      & (sec.)  & Training Dependency \\ \midrule
		\multicolumn{10}{@{}l}{\textbf{CNN - Stereo-with-LiDAR}}                                                                                                                                                                   \\
		PL: AVOD~\cite{Wang_PseudoLiDARVisualDepth_2019}              & 34.05            & 54.53   & 28.25   &  &   ---              &   ---      &   ---     & 0.40    & LiDAR point cloud                                     \\
		OCStereo~\cite{Pon_ObjectcentricStereoMatching_2020}          & 37.60            & 55.15   & 30.25   &  & 17.58              & 24.48      & 15.60     & 0.35    & LiDAR point cloud, instance segmentation              \\
		ZoomNet~\cite{Xu_ZoomNetPartawareAdaptive_2020}               & 38.64            & 55.98   & 30.97   &  &   ---              &   ---      &   ---     & 0.35    & LiDAR point cloud, instance segmentation              \\
		PL++: P-RCNN~\cite{You_PseudoLiDARAccurateDepth_2019}         & 42.43            & 61.11   & 36.99   &  &   ---              &   ---      &   ---     & 0.40    & LiDAR point cloud                                     \\
		E2E-PL: P-RCNN~\cite{Qian_EndtoendPseudoLiDARImagebased_2020} & 43.90            & 64.80   & 38.10   &  &   ---              &   ---      &   ---     & 0.40    & LiDAR point cloud                                     \\
		CDN: PL++~\cite{Garg_WassersteinDistancesStereo_2020}         & 44.86            & 64.31   & 38.11   &  &   ---              &   ---      &   ---     & 0.40    & LiDAR point cloud                                     \\
		Disp R-CNN (velo)~\cite{Sun_DispRCNNStereo_2020}              & 45.78            & 68.21   & 37.73   &  & 25.80              & 37.12      & 22.04     & 0.42    & LiDAR point cloud, instance segmentation, 3D models   \\
		DSGN~\cite{Chen_DSGNDeepStereo_2020}                          & 52.18            & 73.50   & 45.14   &  &   ---              &   ---      &   ---     & 0.67    & LiDAR point cloud                                     \\
		CGStereo~\cite{Li_ConfidenceGuidedStereo_2020}                & 53.58            & 74.39   & 46.50   &  & 24.31              & 33.22      & 20.95     & 0.57    & LiDAR point cloud, semantic segmentation              \\
		CDN: DSGN~\cite{Garg_WassersteinDistancesStereo_2020}         & 54.22            & 74.52   & 46.36   &  &   ---              &   ---      &   ---     & 0.60    & LiDAR point cloud                                     \\
		LIGA-Stereo~\cite{Guo_LIGAStereoLearningLiDAR_2021}           & 64.66            & 81.39   & 57.22   &  & 30.00              & 40.46      & 27.07     & 0.35    & LiDAR point cloud                                     \\ \midrule
		\multicolumn{10}{@{}l}{\textbf{CNN - Stereo-without-LiDAR}}                                                                                                                                                                \\
		TLNet~\cite{Qin_TriangulationLearningNetwork_2019}            & 4.37             & 7.64    & 3.74    &  &   ---              &   ---      &   ---     & ---     & ---                                                   \\
		RT3DStereo~\cite{Konigshof_Realtime3DObject_2019}             & 23.28            & 29.90   & 18.96   &  &  2.45              &  3.28      &  2.35     & 0.08    & Pseudo disparity GT, semantic segmentation            \\
		Stereo R-CNN~\cite{Li_StereoRCNNBased_2019}                   & 30.23            & 47.58   & 23.72   &  &   ---              &   ---      &   ---     & 0.30    & ---                                                   \\
		RTS3D~\cite{Li_RTS3DRealtimeStereo_2021}                      & 37.38            & 58.51   & 31.12   &  &   ---              &   ---      &   ---     & 0.04    & ---                                                   \\
		YOLOStereo3D~\cite{Liu_YOLOStereo3DStepBack_2021}             & 41.25            & 65.68   & 30.42   &  & 19.75              & 28.49      & 16.48     & 0.08    & Pseudo disparity GT                                   \\
		Disp R-CNN~\cite{Sun_DispRCNNStereo_2020}                     & 43.27            & 67.02   & 36.43   &  & 25.40              & 35.75      & 21.79     & 0.42    & Pseudo disparity GT, instance segmentation, 3D models \\ \midrule
		\multicolumn{10}{@{}l}{\textbf{Transformer - Stereo-without-LiDAR}}                                                                                                                                                        \\
		DETR3D~\cite{Wang_DETR3D3DObject_2021}-Binocular              & 17.61            & 29.95   & 13.69   &  &   ---              &   ---      &   ---     & ---     & ---                                                   \\
		TS3D (ours)                                                   & 41.29            & 64.61   & 30.68   &  & 19.56              & 29.17      & 17.20     & 0.09    & Pseudo disparity GT                                   \\ \bottomrule
		\end{tabular}%
\end{table*}

As the first public Transformer-based binocular 3D object detector in the literature at the time of writing, TS3D significantly diminishes the performance gap between CNN-based and Transformer-based detectors. %
In the KITTI test set, the 3D Car and Pedestrian detection AP of TS3D respectively reaches 41.29\% and 19.56\%, which is competitive with that of the latest existing methods without LiDAR-based stereo matching supervision, \textit{i.e.}, YOLOStereo3D~\cite{Liu_YOLOStereo3DStepBack_2021}.
Compared with other models in the table, TS3D is based on the Transformer model. %
Compared with the first part of the table, where LiDAR-based stereo matching supervisions are used during training, TS3D does not rely on LiDAR to provide disparity annotations during training. %
Compared with other detectors that do not rely on LiDAR in the second part, in addition to the difference in the neural network model, the method of this paper has the following differences. %
From the perspective of multi-scale features, TLNet~\cite{Qin_TriangulationLearningNetwork_2019} and RT3DStereo~\cite{Konigshof_Realtime3DObject_2019} use single-scale feature maps; %
Whereas TS3D proposes SPFPN to generate multi-scale feature maps. %
Stereo R-CNN~\cite{Li_StereoRCNNBased_2019} only extracts multi-scale unary features, and uses single-scale features in the 3D object RoI head; %
TS3D uses SPFPN modules in stereo features, and extracts unary and stereo two pyramids of multi-scale features. %
RTS3D~\cite{Li_RTS3DRealtimeStereo_2021}, YOLOStereo3D, and Disp R-CNN~\cite{Sun_DispRCNNStereo_2020} also use the feature pyramid for both unary features and stereo features; %
but do not use a method similar to the SPFPN to fuse multi-scale stereo features. %
From the perspective of additional information required for training, in addition to 3D object annotations, TS3D requires a stereo matching algorithm to generate pseudo-ground-truth disparity maps; %
RT3DStereo relies on semantic segmentation annotations; %
RTS3D takes coarse 3D bounding boxes as input and gradually optimizes the detection results to achieve high-quality 3D object detection; %
Disp R-CNN needs to use external 3D models to generate more accurate pseudo point clouds for candidate objects. %
From the perspective of detection pipelines, Stereo R-CNN requires complex post-processing steps to obtain a more accurate 3D object bounding box after predicting the 3D bounding box; %
TS3D only uses 2D NMS~\cite{Ren_FasterRCNNRealtime_2015} as the post-processing step. %
Stereo R-CNN, TLNet, and Disp R-CNN are two-stage algorithms~\cite{Ren_FasterRCNNRealtime_2015}; %
whereas TS3D is a single-stage detector~\cite{Liu_SSDSingleShot_2016}. %

\cref{tab:ts3d:exp:result} also lists the external dependencies and runtime of the existing methods that do not rely on LiDAR-based stereo matching supervisions.
Among them, Disp R-CNN is higher than TS3D in Moderate 3D Car detection AP, however, its detection speed is slower, respectively 0.42 seconds and 0.67 seconds. %
TS3D takes only 0.09 seconds to detect from a pair of binocular images. %
In addition, TS3D merely relies on the pseudo-ground-truth of stereo matching, while Disp R-CNN requires instance segmentation supervision and 3D models during training; %
The peak GPU memory footprint of our TS3D during inference is 3.1 GB, which is memory efficient compared with the 6.0 GB required by DSGN and 4.9 GB by LIGA-Stereo.

We also visualize three detection results of TS3D on the KITTI validation set (\cref{fig:ts3d:results:success}, one column each). %
The first row visualizes the inverse projection of the disparity map predicted by TS3D and detected 3D bounding boxes (pink); %
The second row illustrates the projection of detected 3D bounding boxes onto the left image. %
The last row shows the projection of 3D ground-truth boxes on the left image. %
In the first column in \cref{fig:ts3d:results:success}, the shadow of trees on the side of the road is a challenge for image-based object detectors, and the distant vehicle ahead is of a relatively small scale. %
Thanks to the explicitly embedded rich details and correspondence information in multi-scale stereo features and Transformer decoders, TS3D can alleviate the problems of scale variation and shadow defects. %
The second column shows a roadside parking scene on a narrow road. %
In addition to the detected roadside cars of various scales, TS3D is also able to detect the car parked vertically on the left. %
The third column showcases the detection of distant cars. %
All four cars are detected, however, the right two cars are of erroneous orientations due to their small scale. %

\section{Conclusions}
\label{sec:ts3d:conclusion}

Transformers are making progress in various fields of computer vision and are expected to be compatible with binocular 3D object detection. %
We have proposed a Transformer-based Stereo-aware 3D object detector (TS3D), which has successfully applied the Transformer model to the binocular 3D object detection task by designing Disparity-Aware Positional Encoding (DAPE) and Stereo Preserving Feature Pyramid Network (SPFPN). %
The image correspondence and 2D location information have been explicitly encoded into object queries by DAPE, providing the decoder with the spatial information of the 3D scene. %
The SPFPN has been designed to provide the Transformer decoder with enriched multi-scale features and object embeddings. %
In the fusion process, SPFPN has preserved the image correspondence information in the stereo features; %
In the cross-scale aggregation, the complementary image correspondence information across scales has been aggregated, and eventually, the stereo feature pyramid with enriched details and correspondence information has been obtained. %

Experiments on the KITTI dataset have shown that TS3D is an effective Transformer model for binocular 3D object detection, which has been competitive with advanced methods of CNN-based models, achieving 41.29\% of the test set Moderate Car detection AP.
Ablation experiments have demonstrated the effectiveness of the proposed DAPE and SPFPN. %
The TS3D could hopefully serve as a baseline for future research on the Transformer model for binocular 3D object detection. %

\bibliographystyle{IEEEtran}
\bibliography{zotero}

\begin{IEEEbiography}[{\includegraphics[width=1in,height=1.25in,clip,keepaspectratio]{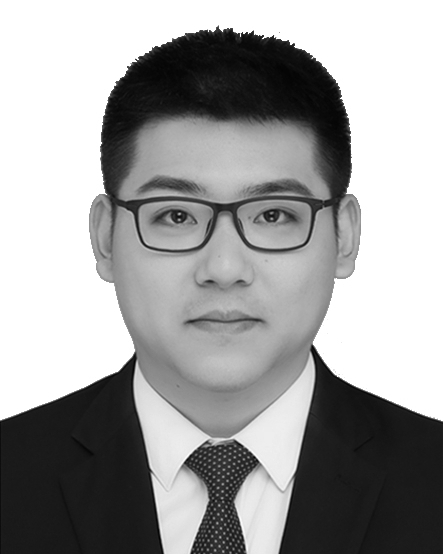}}]{Hanqing Sun}
	received his Ph.D. degree in Information and Communication Engineering from Tianjin University, Tianjin, China, in 2023.
	He is currently an engineer at the Changchun Institute of Optics, Fine Mechanics and Physics, Chinese Academy of Sciences, Changchun, China.
	His research interests include deep learning for stereo vision, remote sensing, and object recognition.
\end{IEEEbiography}
\begin{IEEEbiography}[{\includegraphics[width=1in,height=1.25in,clip,keepaspectratio]{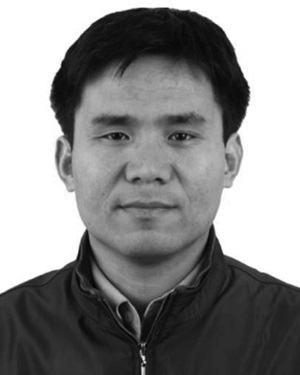}}]{Yanwei Pang}
	(Senior Member, IEEE) received his Ph.D. degree in electronic engineering from the University of Science and Technology of China in 2004.
	He is currently a Professor with Tianjin University, China, where he is also the Founding Director of the Tianjin Key Laboratory of Brain Inspired Intelligence Technology (BIIT).
	His research interests include object detection and image recognition, in which he has published 150 scientific papers, including 40 IEEE Transactions and 30 top conferences (\textit{e.g.}, CVPR, ICCV, and ECCV) papers.
\end{IEEEbiography}
\begin{IEEEbiography}[{\includegraphics[width=1in,height=1.25in,clip,keepaspectratio]{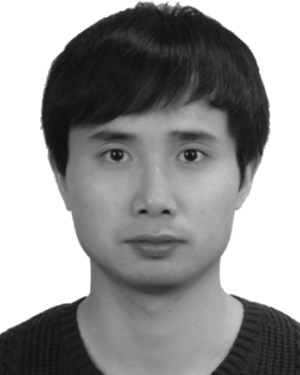}}]{Jiale Cao}
	received his Ph.D. degree in information and communication engineering from Tianjin University, Tianjin, China, in 2018.
	He is currently an Associate Professor with Tianjin University.
	His research interests include object detection and image analysis, in which he has published 20+ IEEE Transactions and CVPR/ICCV/ECCV articles.
	He serves as a regular Program Committee Member for leading computer vision and artificial intelligence conferences, such as CVPR, ICCV, and ECCV.
\end{IEEEbiography}
\begin{IEEEbiography}[{\includegraphics[width=1in,height=1.25in,clip,keepaspectratio]{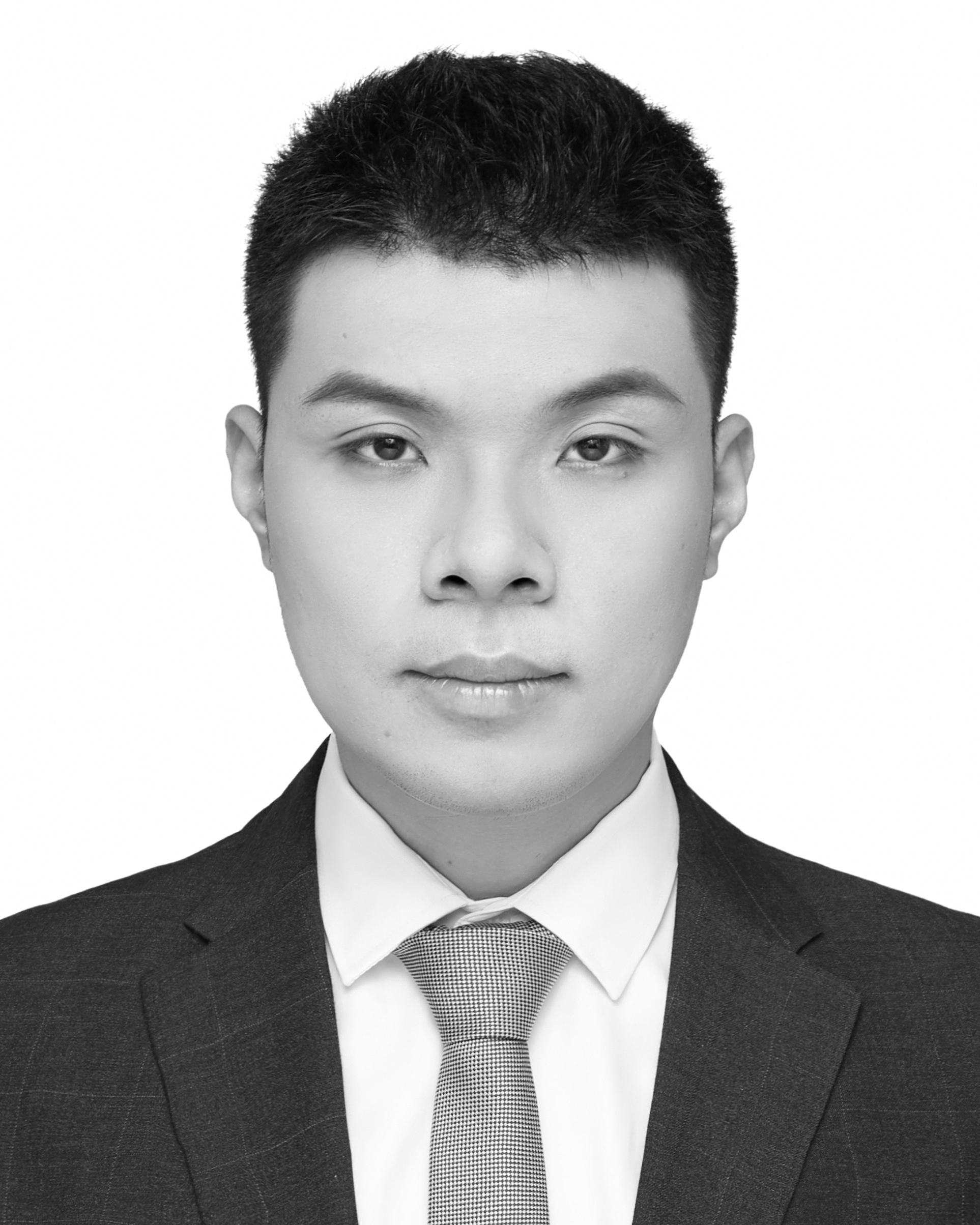}}]{Jin Xie}
	received the Ph.D degree in information and communication engineering from Tianjin University, Tianjin, China, in 2021.
	He is currently an associate professor at the Chongqing University.
	His research interests include machine learning and computer vision, in which he has published 20 papers in CVPR, ICCV, ECCV, IEEE TPAMI, IEEE TIP, IEEE TCSVT, and IEEE TCYB.
\end{IEEEbiography}
\begin{IEEEbiography}[{\includegraphics[width=1in,height=1.25in,clip,keepaspectratio]{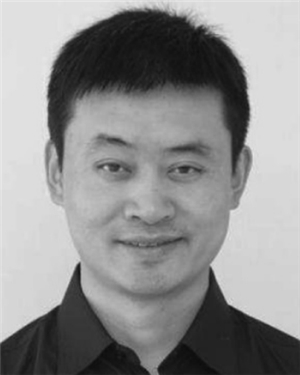}}]{Xuelong Li}
	(Fellow, IEEE) is currently a Full Professor with the School of Artificial Intelligence, Optics and Electronics (iOPEN), Northwestern Polytechnical University, Xi'an, China.
\end{IEEEbiography}

\vfill

\end{document}